\documentclass[a4paper]{llncs}
                              \institute{ }
\titlerunning{Mysterious Machines}
\usepackage{breakcites}
\usepackage{paralist}
\usepackage{array}
\usepackage{graphicx}
\usepackage{listings}

\author{J.-M. Chauvet}
\date{\today}
\title{Prof. Schönhage's Mysterious Machines}
\begin{document}

\maketitle
\begin{abstract}
We give a simple Schönhage's Storage Modification Machine that simulates one iteration of the Rule 110 cellular automaton. This provides an alternative construction to the original Schönhage's proof of the Turing completeness of the eponymous machines. 
\end{abstract}

\section{Introduction}
\label{sec:orgf758b6b}
By a simple construction it is shown that iterations performed by the Rule 110 elementary cellular automaton can be duplicated by a small size Schönhage \emph{Storage Modification Machine}.

\subsection{The Rule 110 cellular automaton}
\label{sec:org7a2fc86}
Rule 110 is one of the elementary cellular automaton rules introduced by Stephen Wolfram in 1983 \cite{Wolfram2002}. It specifies the next color in a cell, white or black, depending on its color and its immediate neighbors. Its rule outcomes are encoded in the binary representation: \(110_{decimal} = 01101110_{binary}\).  The rule 110 cellular automaton is universal, as first conjectured by Wolfram and subsequently proven by Wolfram and Cook \cite{Cook2004}.

\begin{figure}[htbp]
\centering
\includegraphics[width=.9\linewidth]{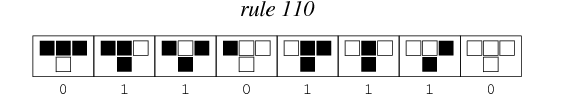}
\caption{Compact representation of the ECA Rule 110}
\end{figure}

Simulation of small universal Turing machines, or other simple universal models such as Post’s tag systems and the cellular automaton Rule 110, is by now a standard way to prove that a large number of other models of computation, including a variety of physically-inspired systems, are computationally universal. In the following, we consider a slightly revised version of Schönhage's Storage Modification Machine (SMM) and propose such a simulation of the Rule 110 automaton.
\newpage

\subsection{Schönhage's Storage Modification Machines}
\label{sec:org1d5be5c}
The variant presented here is from \cite{Guerraoui2009} where it is used to implement population protocol models. A SMM represents a single computing agent. Its memory stores a finite directed graph of equal out-degree nodes \cite{Schoenhage1980}, with a distinguished node called the \emph{center}. Edges of the graph are sometimes called \emph{pointers}. Edges out of each node are labelled by distinct \emph{directions} drawn from a finite set \(D\).

Any string \(x \in D^*\) refers to the node \(p(x)\) reached from the center by following the sequence of directions labelled by \(x\). In the variant used here, nodes may have different out-degrees, and we set \(p(x) = \emptyset\) when \(x\) is not a valid path in the graph.

SMMs are additionally characterized by a \emph{program} which is a finite list of consecutively numbered instructions. The basic instruction set is as follows:

\begin{itemize}
\item \emph{\textbf{new} label} creates a new labelled node and makes it the center, setting all its outgoing edges to the previous center.
\item \emph{\textbf{set} xd \textbf{to} y} where \(x,y\) are paths in \(D^*\) and \(d \in D\) is a direction, redirects the \(d\) edge of \(p(x)\) to point to \(p(y)\).
\item \emph{\textbf{center} x} where \(x\) is a path, moves the center to \(p(x)\).
\item \emph{\textbf{if} x y \textbf{then} ln} where \(x,y\) are paths and \emph{ln} a line number, jumps to line \emph{ln} if \(p(x) = p(y)\) and skips to the next line if not. Line numbers can be absolute, \emph{ln}, or relative to the current line number, \emph{+ln} or \emph{-ln}.
\item \emph{\textbf{stop} message} halts the SMM, printing \emph{message}.
\end{itemize}

Other instructions are easily thought of for adding some form of I/O capabilities to the base SMM. A simple measure of a SMM space complexity is the number of reachable nodes (from the center) at any one time during execution. Van Emde Boas \cite{VANEMDEBOAS1989103} has shown that a SMM can simulate a Turing Machine.
\newpage

\section{A Size 4 SMM Simulating Rule 110}
\label{sec:org309128f}
The simple construction maps a cell of the Rule 110 automaton to a node of the SMM. A row of cells is a doubly-linked chain of nodes and as each row represents a new generation, or the result of an iteration of the cellular automaton, we make each node point to its previous generation in the predecessor row. Finally, the cell state, on (black) or off (white), is captured conventionally by a fourth outgoing edge either to self, for an on-state, or to another node, for an off-state. The out-degree of nodes is then 4 and the directions are aptly named \(D = {n, s, e, w}\), the \(n\) direction being used to indicate the cell state and \(s\) as a pointer to the predecessor node during SMM operations.

With these specifications, the initial row of the cellular automaton made of one central on-cell flanked by off-cells is obtained by running the \texttt{init} program on the size 4 SMM:

\begin{center}
  \begin{tabular}[htbp]{ m{5cm} m{5cm} }

\begin{lstlisting}[caption={The \texttt{init} program builds an initial row of 7 cells.}]
 1 new center-T0
 2 new right1-T0
 3 set n
 4 set we
 5 new right2-T0
 6 set n
 7 set we
 8 new right3-T0
 9 set n
10 set we
11 ctr www
12 new left1-T0
13 set n
14 set ew
15 new left2-T0
16 set n
17 set ew
18 new left3-T0
19 set n
20 set ew
21 set eeeeeee
22 set w eeeeee
23 stop
\end{lstlisting}

&

\includegraphics[height=9cm]{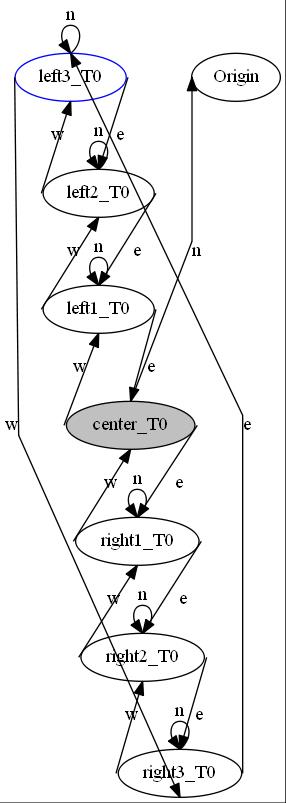}

  \end{tabular}
\end{center}

The graph is grown from its center node, pointing originally to the default \emph{Origin} center, first on the right, which appears as the top half of the column, re-centered and finally to the left, which appears as the bottom half of the column in Figure 2.

A single iteration of the Rule 110 is captured by another SMM program, \texttt{iterate} which is repeatedly executed to simulate consecutive iterations of the cellular automaton.

\begin{lstlisting}[caption={The \texttt{iterate} program implements Rule 110}]
 1 set ns
 2 if swn sw +2
 3 if s s +7
 4 if sn s +2
 5 if s s +4
 6 if sen se +2
 7 if s s +2
 8 set n
 9 if s s +7
10 if sn s  +4
11 if sen se +2
12 set n
13 if s s +3
14 if sen se -2
15 if s s -2
\end{lstlisting}

The previous block of instructions is executed, at creation time for each node in the new generation row. The \(n\) directions of the three nodes in the neighborhood are tested for their returning to the same themselves or not (lines \(2\), \(4\), \(6\), \(10\) and \(11\)), and  and the pattern of jumps in the \emph{if} statements determines the \(n\) direction (lines \(8\) and \(12\)) of the current node\footnote{The block of instructions presented is certainly not optimized for size!}.

Should we need to implement a loop over rows of the ECA in the restricted SMM assembly-like language, we would add two ancillary directions, say \emph{B} and \emph{T}, to each node. The first one, \emph{B},  would always point to the \emph{Origin} node, while \emph{T} would point to the parent node in the previous row. Stopping after \emph{max} iterations is then controlled by an \texttt{if TT...T B then <stop line number>} where \emph{T} is repeated \emph{max} times in the path. Similarly, iteration on cells within a single row would cost two more directions, the first one constantly pointing back to the first node in the row and the second one to the predecessor in the row.

After 9 iterations, simulation of the Rule 110 on an initial row of 21 cells produces the storage graph in Figure 2.

\begin{figure}[htbp]
\centering
\includegraphics[height=8cm]{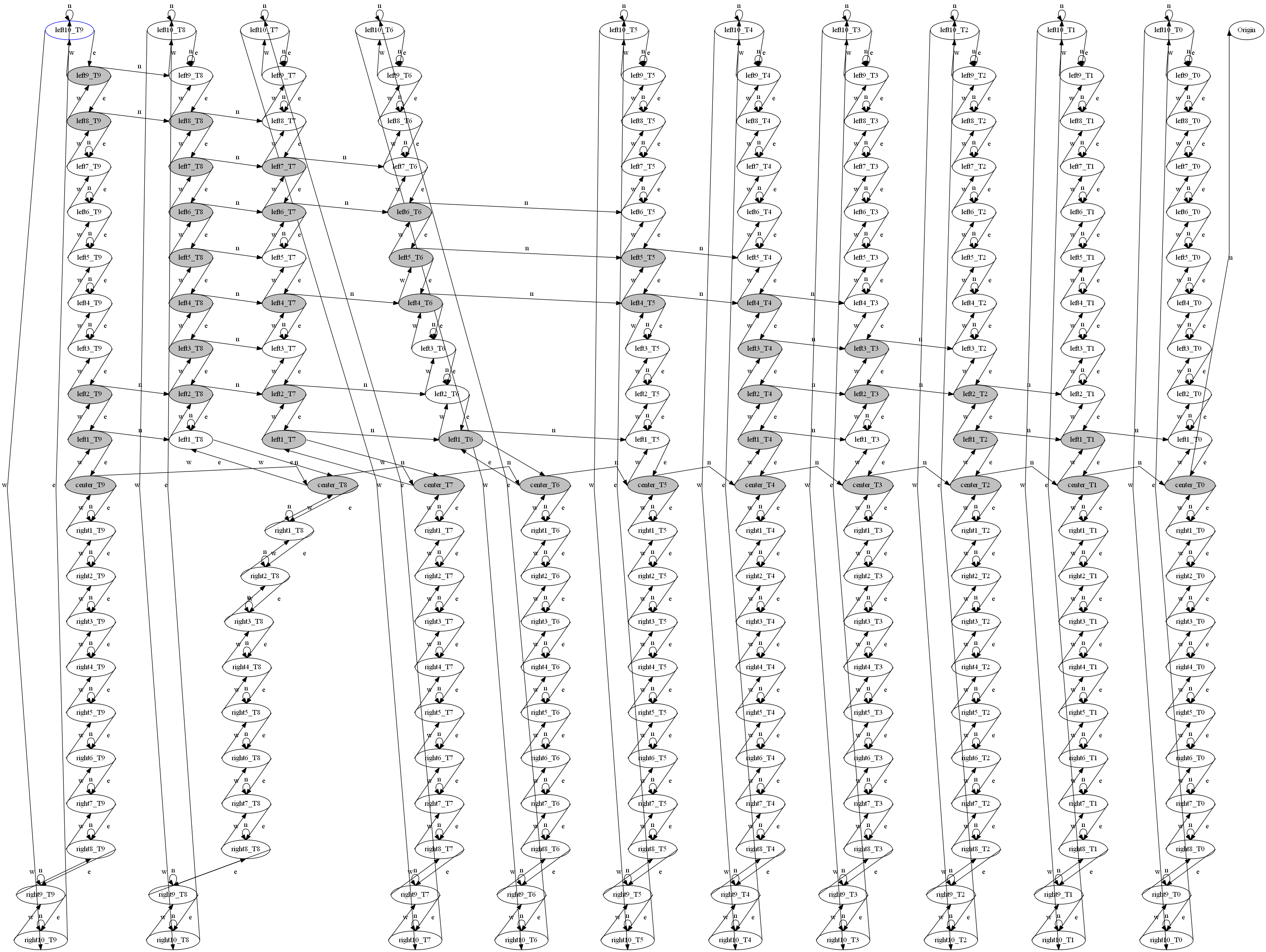}
\caption{SMM simulation of Rule 110 after 9 iterations on a 21-cell row.}
\end{figure}

Clearly the space complexity of this construction is \(R*I\) where \(R\) is the number of cells in a row, and \(I\) is the number of iterations.

\section{Conclusion}
\label{sec:org2b1484b}
It is well known that Schönhage's Storage Modification Machines can simulate Turing Machines. This paper provides a construction showing that they can simulate the famous Rule 110 elementary cellular automaton. Along the same principles, an out-degree-4 SMM can simulate any of the 256 ECA. Related questions on the minimal (space) complexity SMM required for simulation of more complex cellular automata may be addressed by looking into optimizing this construction.

\bibliographystyle{plain}
\bibliography{smm}

\begin{thebibliography}{1}

\bibitem{Cook2004}
Matthew Cook et~al.
\newblock {Universality in elementary cellular automata}.
\newblock {\em Complex systems}, 15(1):1--40, 2004.

\bibitem{Guerraoui2009}
Rachid Guerraoui and Eric Ruppert.
\newblock {Names Trump Malice: Tiny Mobile Agents Can Tolerate Byzantine
  Failures}.
\newblock In {\em Proceedings of the 36th Internatilonal Collogquium on
  Automata, Languages and Programming: Part II}, ICALP '09, page 484–495,
  Berlin, Heidelberg, 2009. Springer-Verlag.

\bibitem{Schoenhage1980}
A.~Schönhage.
\newblock {Storage Modification Machines}.
\newblock {\em {SIAM Journal on Computing}}, 9(3):490--508, 1980.

\bibitem{VANEMDEBOAS1989103}
Peter {van Emde Boas}.
\newblock Space measures for storage modification machines.
\newblock {\em Information Processing Letters}, 30(2):103--110, 1989.

\bibitem{Wolfram2002}
Stephen Wolfram.
\newblock {\em A New Kind of Science}.
\newblock Wolfram Media, 2002.

\end{thebibliography}
\end{document}